\documentclass[runningheads]{llncs}
\usepackage{graphicx}
\usepackage{amsmath,amssymb} 
\usepackage{color}
\usepackage{cite}
\usepackage{subfigure} 
\usepackage{multirow}
\usepackage{array}
\usepackage[table,xcdraw]{xcolor}

\begin{document}

\title{GD-GAN: Generative Adversarial Networks for Trajectory Prediction and Group Detection in Crowds} 
\titlerunning{GD-GAN} 


\author{Tharindu Fernando\inst{1} \and
Simon Denman\inst{1}\and
Sridha Sridharan\inst{1} \and Clinton Fookes \inst{1}}
%

\authorrunning{T. Fernando et al.} 


\institute{ Image and Video Research Laboratory, SAIVT, Queensland University of Technology (QUT), Australia. \\
\email{\{t.warnakulasuriya, s.denman, s.sridharan, c.fookes\}@qut.edu.au}}

\maketitle

\begin{abstract}
This paper presents a novel deep learning framework for human trajectory prediction and detecting social group membership in crowds. We introduce a generative adversarial pipeline which preserves the spatio-temporal structure of the pedestrian's neighbourhood, enabling us to extract relevant attributes describing their social identity. We formulate the group detection task as an unsupervised learning problem, obviating the need for supervised learning of group memberships via hand labeled databases, allowing us to directly employ the proposed framework in different surveillance settings. We evaluate the proposed trajectory prediction and group detection frameworks on multiple public benchmarks, and for both tasks the proposed method demonstrates its capability to better anticipate human sociological behaviour compared to the existing state-of-the-art methods. \footnote{This research was supported by the Australian Research Council's Linkage Project LP140100282 ``Improving Productivity and Efficiency of Australian Airports''}

\keywords{Group detection  \and Generative Adversarial Networks \and Trajectory Prediction.}
\end{abstract}
\section{Introduction}
\label{sec:intro}
Understanding and predicting crowd behaviour plays a pivotal role in video based surveillance; and as such is becoming essential for discovering public safety risks, and predicting crimes or patterns of interest. Recently, focus has been given to understanding human behaviour at a group level, leveraging observed social interactions. Researchers have shown this to be important as interactions occur at a group level, rather than at an individual or whole of crowd level. 

As such we believe group detection has become a mandatory part of an intelligent surveillance system; however this group detection task presents several new challenges ~\cite{solera2016socially,solera2013structured}. Other than identifying and tracking pedestrians from video, modelling the semantics of human social interaction and cultural gestures over a short sequence of clips is extremely challenging. Several attempts \cite{solera2016socially,solera2013structured,pellegrini2010improving,yamaguchi2011you} have been made to incorporate handcrafted  physics based features such as relative distance between pedestrians, trajectory shape and motion based features to model their social affinity. Hall et. al \cite{hall1966hidden} proposed a proxemic theory for such physical interactions based on different distance boundaries; however recent works \cite{solera2016socially,solera2013structured} have shown these quantisations fail in cluttered environments. 

Furthermore, proximity doesn't always describe the group membership. For instance two pedestrians sharing a common goal may start their trajectories in two distinct source positions, however, meet in the middle. Hence we believe being reliant on a handful of handcrafted features to be sub-optimal \cite{isola2017image,alahi2016social,fernando2018task}. 

\begin{figure}[htbp]
  \centering
    \includegraphics[width=\linewidth]{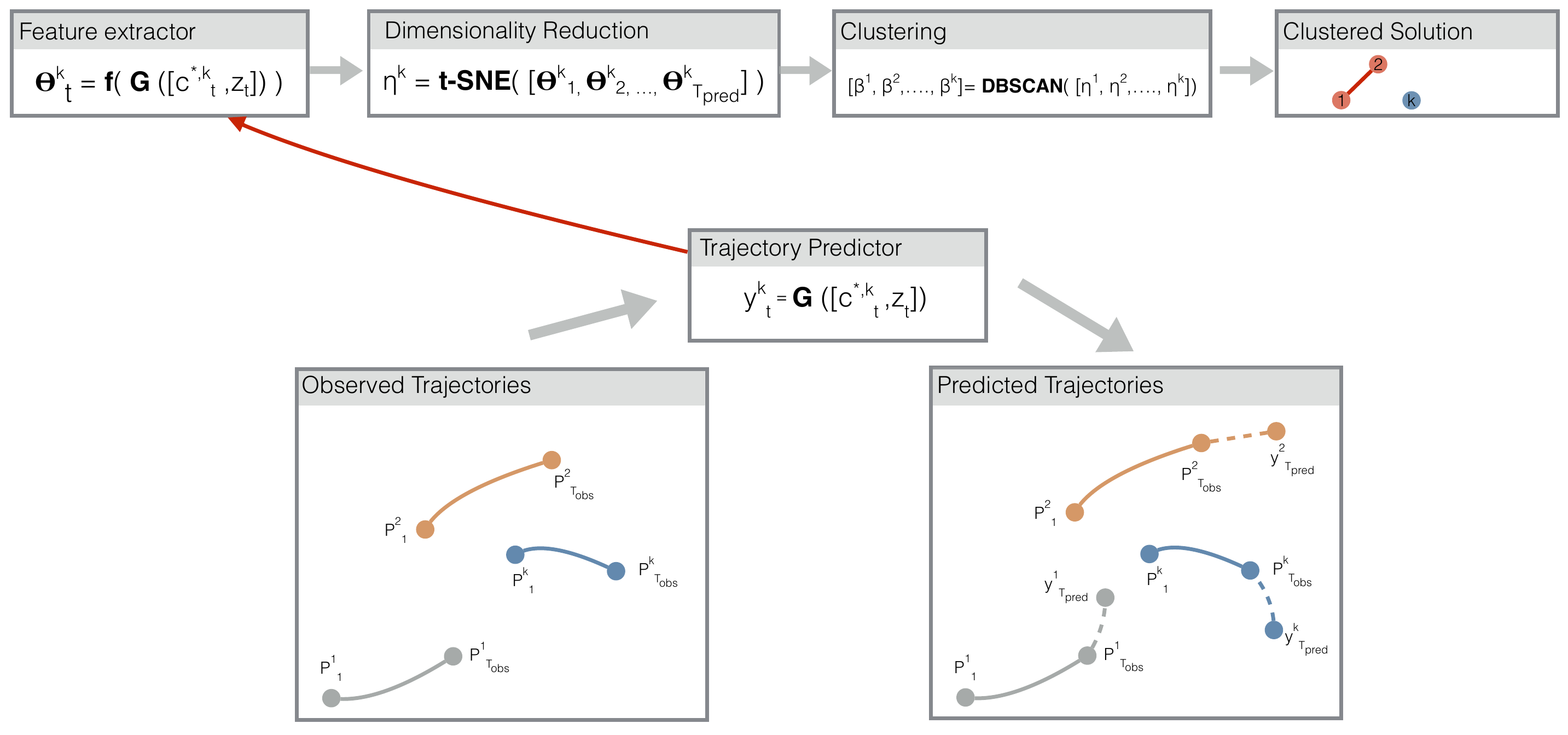}
   \caption{Proposed group detection framework:  After observing short segments of trajectories for each pedestrian in the scene, we apply the proposed trajectory prediction algorithm to forecast their future trajectories. The context representation generated at this step is extracted and compressed using t-SNE dimensionality reduction. Finally, the DBSCAN clustering algorithm is applied to detect the pedestrian groups.}
  \label{fig:model}
\end{figure}

To this end we propose a deep learning algorithm which automatically learns these group attributes. We take inspiration from the trajectory modelling approaches of \cite{fernando2017soft+} and \cite{fernando2018tracking}, where the approaches capture contextual information from the local neighbourhood.  We further augment this approach with a Generative Adversarial Network (GAN) \cite{gupta2018social,sadeghian2018sophie,fernando2018task} learning pipeline where we learn a custom, task specific loss function which is specifically tailored for future trajectory prediction, learning to imitate complex human behaviours. 

Fig. \ref{fig:model} illustrates the proposed approach. First, we observe short segments of trajectories from 1 to $T_{obs}$ for each pedestrian, $p^{k}$, in the scene. Then, we apply the proposed trajectory prediction algorithm to forecast their future trajectories from $T_{obs +1} -T_{pred}$. This step generates hidden context representations for each pedestrian describing the current environmental context in the local neighbourhood of the pedestrian. We then apply t-SNE dimensionality reduction to extract the most discriminative features, and we detect the pedestrian groups by clustering these reduced features. 

The simplistic nature of the proposed framework offers direct transferability among different environments when compared to the supervised learning approaches of\cite{solera2016socially,solera2013structured,pellegrini2010improving,yamaguchi2011you}, which require re-training of the group detection process whenever the surveillance scene changes. This ability is a result of the proposed deep feature learning framework which learns the required group attributes automatically and attains commendable results among the state-of-the-art. 

Novel contributions of this paper can be summarised as follows:

\begin{itemize}
\item We propose a novel GAN pipeline which jointly learns informative latent features for pedestrian trajectory forecasting and group detection. 
\item We remove the supervised learning requirement for group detection, allowing direct transferability among different surveillance scenes. 
\item We demonstrate how the original GAN objective could be augmented with sparsity regularisation to learn powerful features which are informative to both trajectory forecasting and group detection tasks. 
\item We provide extensive evaluations of the proposed method on multiple public benchmarks where the proposed method is able to generate notable performance, especially among unsupervised learning based methods. 
\item We present visual evidence on how the proposed trajectory modelling scheme has been able to embed social interaction attributes into its encoding scheme. 
\end{itemize}

\section{Related Work}
Related literature is categorised into human behaviour prediction approaches (see Sec. \ref{sec:related_work_human_behaviour_prediction}); and group detection architectures (see Sec. \ref{sec:related_work_group_detection}).

\subsection{Human Behaviour Prediction}
\label{sec:related_work_human_behaviour_prediction}
Social Force models \cite{helbing1995social,yamaguchi2011you}, which rely on the attractive and repulsive forces between pedestrians to model their future behaviour, have been extensively applied for modelling human navigational behaviour. However with the dawn of deep learning, these methods have been replaced as they have been shown to ill represent the structure of human decision making \cite{fernando2017soft+,fernando2018tree,gupta2018social}. 

One of the most popular deep learning methods is the social LSTM \cite{alahi2016social} model which represents the pedestrians in the local neighbourhood using LSTMs and then generates their future trajectory by systematically pooling the relavant information. This removes the need for handcrafted features and learns the required feature vectors automatically through the encoded trajectory representation. This architecture is further augmented in \cite{fernando2017soft+} where the authors propose a more efficient method to embed the local neighbourhood information via a soft and hardwired attention framework. They demonstrate the importance of fully capturing the context information, which includes the short-term history of the pedestrian of interest as well as their neighbours. 

Generative Adversarial Networks (GANs) \cite{gupta2018social,sadeghian2018sophie,fernando2018task} propose a task specific loss function learning process where the training objective is a minmax game between the generative and discriminative models. These methods have shown promising results, overcoming the intractable computation of a loss function, in tasks such as autonomous driving \cite{fernando2018learning,li2017infogail}, saliency prediction \cite{fernando2018task,pan2017salgan}, image to image translation \cite{isola2017image} and human navigation modelling \cite{gupta2018social,sadeghian2018sophie}.

Even though the proposed GAN based trajectory modelling approach exhibits several similarities to recent works in \cite{gupta2018social,sadeghian2018sophie}, the proposed work differs in multiple aspects. Firstly, instead of using CNN features to extract the local structure of the neighbourhood as in \cite{sadeghian2018sophie}, pooling out only the current state of the neighbourhood as in \cite{gupta2018social}, or discarding the available historical behaviour which is shown to be ineffective \cite{fernando2017soft+,fernando2018tree,sadeghian2018sophie}; we propose an efficient method to embed the local neighbourhood context based on the soft and hardwired attention framework proposed in \cite{fernando2017soft+}. Secondly, as we have an additional objective of localising the groups in the given crowd, we propose an augmentation to the original GAN objective which regularises the sparsity of the generator embeddings, generating more discriminative features and aiding the clustering processes. 

\subsection{Group Detection}
\label{sec:related_work_group_detection}

Some earlier works in group detection \cite{cristani2011social,setti2013multi} employ the concept of F-formations \cite{kendon1990conducting}, which can be seen as specific orientation patterns that individuals engage in when in a group. However such methods are only suited to stationary groups. 

In a separate line of work researchers have analysed pedestrian trajectories to detect groups. Pellegrinin et. al \cite{pellegrini2010improving} applied Conditional Random Fields to jointly predict the future trajectory of the pedestrian of interest as well as their group membership. \cite{yamaguchi2011you} utilises distance, speed and overlap time to train a linear SVM to classify whether two pedestrians are in the same group or not. In contrast to these supervised methods, Ge et. al \cite{ge2012vision} proposed using agglomerative clustering of speed and proximity features to extract pedestrian groups.

Most recently Solera et. al \cite{solera2013structured} proposed proximity and occupancy based social features to detect groups using a trained structural SVM. In \cite{solera2016socially} the authors extend this preliminary work with the introduction of sociologically inspired features such as path convergence and trajectory shape. However these supervised learning mechanisms rely on hand labeled datasets to learn group segmentation, limiting the methods applicability. Furthermore, the above methods all utilise a predefined set of handcrafted features to describe the sociological identity of each pedestrian, which may be suboptimal. Motivated by the impressive results obtained in \cite{fernando2017soft+} with the augmented context embedding, we make the first effort to learn  group attributes automatically and jointly through trajectory prediction. 

\section{Architecture}
\subsection{Neighbourhood Modelling}
\label{sec:neighbourhood_modelling}
\begin{figure}[htbp]
  \centering
    \includegraphics[width=.8\linewidth]{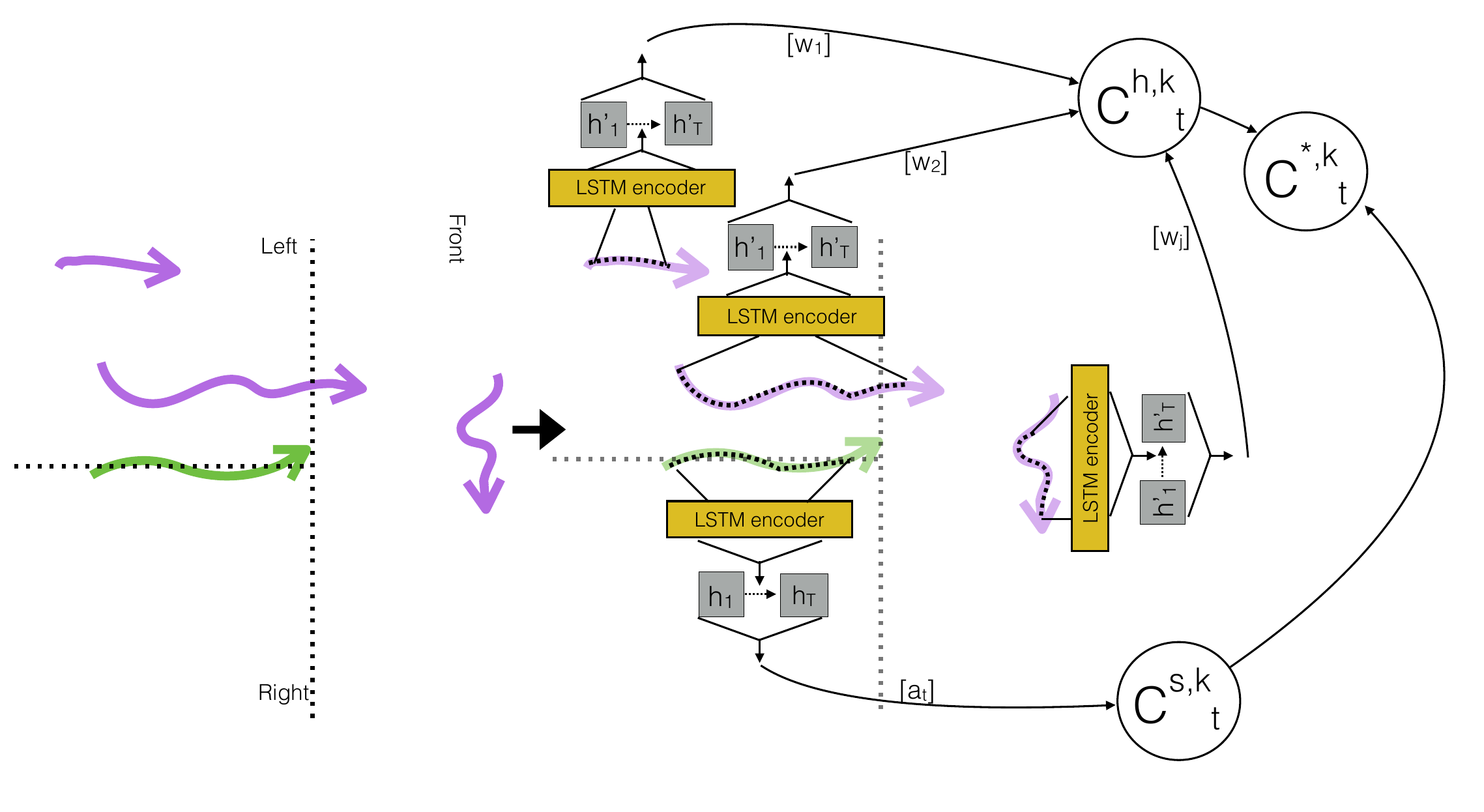}
   \caption{Proposed neighbourhood modelling scheme \cite{fernando2017soft+}:  A sample surveillance scene is shown on the left. The trajectory of the pedestrian of interest, $k$ , is shown in green, and has two neighbours (in purple) to the left, one in front and none on right. The neighbourhood encoding scheme shown on the right: Trajectory information is encoded with LSTM encoders. A soft attention context vector $C^{s,k}_{t}$ is used to embed trajectory information from the pedestrian of interest, and a hardwired attention context vector $C^{h,k}_t$ is used for neighbouring trajectories.  In order to generate $C^{s,k}_{t}$ we use a soft attention function denoted $a_t$ in the above figure, and the hardwired weights are denoted by $w$. The merged context vector $C^{*,k}_{t}$ is then generated by merging $C^{s,k}_{t}$ and $C^{h,k}_t$.}
  \label{fig:neighbourhood_modelling}
\end{figure}

We use the trajectory modelling framework of \cite{fernando2017soft+} (shown in Fig. \ref{fig:neighbourhood_modelling}) for modelling the local neighbourhood of the pedestrian of interest.  

Let the trajectory of the pedestrian $k$, from frame 1 to $T_{obs}$ be given by,
\begin{equation}
p^k = [p_1, \ldots, p_{T_{obs}}],
\end{equation}
where the trajectory is composed of points in a Cartesian grid. Then we pass each trajectory through an LSTM \cite{hochreiter1997long} encoder to generate its hidden embeddings, 
\begin{equation}
h^k_t =  LSTM (p_t^k, h_{t-1}^k),
\label{eq:lstm_encoding}
\end{equation}
generating a sequence of embeddings,
\begin{equation}
h^k =   [h_1^k, \ldots ,h_{T_{obs}}^k].
\end{equation}

Following \cite{fernando2017soft+}, the trajectory of the pedestrian of interest is embedded with soft attention such that, 
\begin{equation}
C^{s,k}_{t}=\sum_{j=1}^{T_{obs}} \alpha_{tj}h^k_j ,
\end{equation}
which is the weighted sum of hidden states. The weight $\alpha_{tj}$ is computed by,
\begin{equation}
\alpha_{tj}=\cfrac{exp(e_{tj})}{\sum_{l=1}^{T} exp(e_{tl})} , 
\end{equation}
\begin{equation}
e_{tj}=a(h^{k}_{t-1},h^k_j).
\end{equation}
The function $a$ is a feed forward neural network jointly trained with the other components. 

To embed the effect of the neighbouring trajectories we use the hardwired attention context vector $C^{h,k}_t$ from \cite{fernando2017soft+}. The hardwired weight $w$ is computed by, 
\begin{equation}
w^n_{j}=\cfrac{1}{\mathrm{dist}(n,j)},
\label{eq:hw_weight}
\end{equation}
where $\mathrm{dist}(n,j)$ is the distance between the $n^{th}$ neighbour and the pedestrian of interest at the $j^{th}$ time instant. Then we compute $C^{h,k}_t$ as the aggregation for all the neighbours such that,
\begin{equation}
C^{h,k}_t=\sum_{n=1}^{N}\sum_{j=1}^{T_{obs}} w^n_{j}h^{n}_{j} ,
\end{equation}
where there are $N$ neighbouring trajectories in the local neighbourhood, and $h^{n}_{j}$ is the encoded hidden state of the $n^{th}$ neighbour at the $j^{th}$ time instant. Finally we merge the soft attention and hardwired attention context vectors to represent the current neighbourhood context such that, 
\begin{equation}
C_{t}^{*,k}=\mathrm{tanh}([C^{s,k}_{t}, C^{h,k}]).
\label{eq:context_vector}
\end{equation}

\subsection{Trajectory Prediction}
\label{sec:trajectory_gen}
Unlike \cite{fernando2017soft+}, we use a GAN to predict the future trajectory. There exists a minmax game between the generator (G) and the discriminator (D) guiding the model G to be closer to the ground truth distribution. The process is guided by learning a custom loss function which generates an additional advantage when modelling complex behaviours such as human navigation, where multiple factors such as human preferences and sociological factors influence behaviour. 

Trajectory prediction can be formulated as observing the trajectory from time 1 to $T_{obs}$, denoted as $[p_1, \ldots, p_{T_{obs}}]$, and forecasting the future trajectory for time $T_{obs+1}$ to $T_{pred}$, denoted as $[y_{T_{obs+1}}, \ldots, y_{T_{pred}}]$. The GAN learns a mapping from a noise vector $z$ to an output vector y, $G: z \rightarrow y$ \cite{fernando2018task}. Adding the notion of time, the output of the model $y_t$ can be written as $G: z_t \rightarrow y_t$.

We augment the generic GAN mapping to be conditional on the current neighbourhood context $C_{t}^{*}$, $G: ({C_{t}^{*}, z_t}) \rightarrow y_t$, such that the synthesised trajectories follow the social navigational rules that are dictated by the environment. 

This objective can be written as,
\begin{equation}
V=\mathbb{E}_{y_t, C^*_t \sim p_{data}}([ log D(C^*_t, y_t)]) + \mathbb{E}_{C^*_t \sim p_{data}, z_t \sim noise}([ 1- log D(C^*_t, G(C^*_t, z_t))]). 
\label{eq:gan}
\end{equation}
Our final aim is to utilise the hidden state embeddings from the trajectory generator to discover the pedestrian groups via clustering those embeddings. Hence having a sparse feature vector for clustering is beneficial as they are more discriminative compared to their dense counterparts \cite{figueroa2017learning}. Hence we augment the objective in Eq. \ref{eq:gan} with a sparsity regulariser such that,
\begin{equation}
L_1=|| f(G (C^{*}_{t}, z_t)) ||_1 ,
\label{eq:sparsity}
\end{equation}
and 
\begin{equation}
V^*=V + \lambda L_1 ,
\label{eq:gan_augmented}
\end{equation}
where $f$ is a feature extraction function which extracts the hidden embeddings from the trajectory generator $G$, and $\lambda$ is a weight vector which controls the tradeoff between the GAN objective and the sparsity constraint. 

\begin{figure}[htb]
  \centering
    \includegraphics[width=\linewidth]{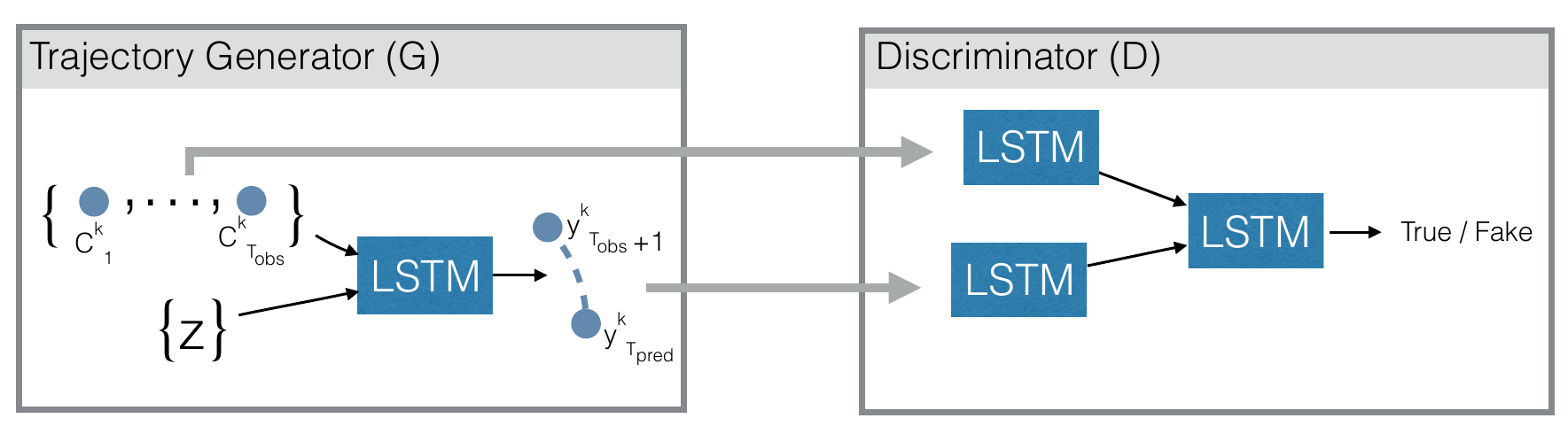}
   \caption{Proposed trajectory prediction framework: The generator model $G$ samples from the noise distribution $z$ and synthesises a trajectory $y_t$, which is conditioned upon the local neighbourhood context $C_t^*$. The discriminator $D$ considers both $y_t$ and $C_t^*$ when classifying the authenticity of the trajectory.} 
  \label{fig:GAN}
\end{figure}

The architecture of the proposed trajectory prediction framework is presented in Fig. \ref{fig:GAN}. We utilise LSTMs as the Generator ($G$) and the Discriminator ($D$) models. $G$ samples from the noise distribution, $z$, and synthesises a trajectory for the pedestrian motion which is conditioned upon the local neighbourhood context, $C_t^*$, of that particular pedestrian. Utilising these predicted trajectories, $y_t$, and the context embeddings, $C_t^*$, $D$ tries to discriminate between the synthesised and ground truth human trajectories. 

\subsection{Group Detection}
Fig. \ref{fig:model} illustrates the proposed group detection framework. We pass each trajectory in the given scene through Eq. \ref{eq:lstm_encoding} to Eq. \ref{eq:context_vector} and generate the neighbourhood embeddings, $C_{t}^{*,k}$. Then using the feature extraction function $f$ we extract the hidden layer activations for each pedestrian $k$ such that, 
\begin{equation}
\theta_{t}^{k}=f(G (C^{*,k}_{t}, z_t)) .
\label{eq:feature_extractor}
\end{equation}

Then we pass the extracted feature vectors through a t-SNE \cite{maaten2008visualizing} dimensionality reduction step. The authors in \cite{figueroa2017learning} have shown that it is inefficient to cluster dense deep features. However they have shown the t-SNE algorithm to generate discriminative features capturing the salient aspects in each feature dimension.  Hence we apply t-SNE for the $k^{th}$ pedestrian in the scene such that, 
\begin{equation}
\eta^{k}=\text{t-SNE}( [\theta^{k}_{1}, \ldots, \theta^{k}_{T_{obs}}]) .
\label{eq:t_SNE}
\end{equation}

As the final step we apply DBSCAN \cite{ester1996density} to discover similar activation patterns, hence segmenting the pedestrian groups. DBSCAN enables us to cluster the data on the fly without specifying the number of clusters. The process can be written as,
\begin{equation}
[\beta^{1}, \ldots, \beta^{N} ]=\mathrm{DBSCAN}( [\eta^{1}, \ldots, \eta^{N} ]) ,
\label{eq:DBSCAN}
\end{equation}
where there are $N$ pedestrians in the given scene and $\beta^n \in [\beta^{1}, \ldots, \beta^{N} ]$ are the generated cluster identities.

\section{Evaluation and Discussion}
\subsection{Implementation Details}
When encoding the neighbourhood information, similar to \cite{fernando2017soft+}, we consider the closest 10 neighbours from each of the left, right, and front directions of the pedestrian of interest. If there are more than 10 neighbours in any direction, we take the closest 9 trajectories and the mean trajectory of the remaining neighbours. If a trajectory has less than 10 neighbours, we created dummy trajectories with hardwired weights (i.e Eq. \ref{eq:hw_weight}) of 0, such that we always have 10 neighbours. 

For all LSTMs, including LSTMs for neighbourhood modelling (i.e Sec. \ref{sec:neighbourhood_modelling}), the trajectory generator and the discriminator (i.e Sec \ref{sec:trajectory_gen}), we use a hidden state embedding size of 300 units. We trained the trajectory prediction framework iteratively, alternating between a generator epoch and a discriminator epoch with the Adam \cite{kingma2014adam} optimiser, using a mini-batch size of 32 and a learning rate of 0.001 for 500 epochs. The hyper parameter $\lambda = 0.2$, and the hyper parameters of DBSCAN, epsilon$=0.50$, minPts$=1$, are chosen experimentally.

\subsection{Evaluation of the Trajectory Prediction}
\label{sec:ped_eval}
\subsubsection{Datasets}
We evaluate the proposed trajectory predictor framework on the publicly available walking pedestrian dataset (BIWI) \cite{pellegrini2009you}, Crowds By Examples (CBE) \cite{lerner2007crowds} dataset and Vittorio Emanuele II  Gallery (VEIIG) dataset \cite{bandini2014towards}. The BIWI dataset records two scenes, one outside a university (ETH) and one at a bus stop (Hotel). CBE records a single video stream with a medium density crowd outside a university (Student 003). The VEIIG dataset provides one video sequence from an overhead camera in the Vittorio Emanuele II Gallery (gall). The training, testing and validation splits for BIWI, CBE and VEIIG are taken from \cite{pellegrini2009you}, \cite{solera2013structured} and \cite{solera2016socially} respectively. 

These datasets include a variety of pedestrian social navigation scenarios including collisions, collision avoidance and group movements, hence presenting challenging settings for evaluation. Compared to BIWI which has low crowd densities, CBE and VEIIG contain higher crowd densities and as a result more challenging crowd behaviour arrangements, continuously varying from medium to high densities.

\subsubsection{Evaluation Metrics}
\label{sec:track_error_metrics}
Similar to \cite{sadeghian2018sophie,gupta2018social} we evaluated the trajectory prediction performance with the following 2 error metrics: Average Displacement Error (ADE) and Final Displacement Error (FDE). Please refer to \cite{sadeghian2018sophie,gupta2018social} for details. 

\subsubsection{Baselines and Evaluation}
We compared our trajectory prediction model to 5 state-of-the-art baselines. As the first baseline we use the Social Force (SF) model introduced in \cite{yamaguchi2011you}, where the destination direction is taken as an input to the model and we train a linear SVM model similar to \cite{fernando2017soft+} to generate this input. We use the Social-LSTM (So-LSTM) model of \cite{alahi2016social} as the next baseline and the neighbourhood size hyper-parameter is set to 32 px. We also compare to the Soft $+$ Hardwired Attention (SHA) model of \cite{fernando2017soft+} and similar to the proposed model we set the embedding dimension to be 300 units and consider a 30 total neighbouring trajectories. We also considered the Social GAN (So-GAN) \cite{gupta2018social} and attentive GAN (SoPhie) \cite{sadeghian2018sophie} models. To provide fair comparisons we set the hidden state dimensions for the encoder and decoder models of So-GAN and SoPhie to be 300 units. For all models we observe the first 15 frames (i.e 1- $T_{obs}$) and predicted the future trajectory for the next 15 frames (i.e $T_{obs+1}$ - $T_{pred}$). 
\begin{table}[htb]
\centering
\caption{Quantitative results for the BIWI \cite{pellegrini2009you}, CBE \cite{lerner2007crowds} and VEIIG \cite{bandini2014towards} datasets. In all methods the forecast trajectories are of length 15 frames. Error metrics are as in Sec. \ref{sec:track_error_metrics}. `-' refers to unavailability of that specific evaluation. The best values are denoted in bold.}
\label{tab:track_prediction}
\resizebox{0.980\textwidth}{!}{
\begin{tabular}{|c|c|c|c|c|c|c|c|}
\hline
          
Metric & Dataset                          & SF \cite{yamaguchi2011you}                            & So-LSTM \cite{alahi2016social}                      & SHA \cite{fernando2017soft+}                          & So-GAN \cite{gupta2018social}  & SoPhie \cite{sadeghian2018sophie}  & Proposed\\ \hline
                         & ETH (BIWI)                                         &        1.42                                         &                1.05                                 &          0.90                                       & 0.92                                            & 0.81        &     \textbf{0.63}                               \\ \cline{2-8} 
                         & Hotel (BIWI)                                       &         1.03                                        &                0.98                                 &         0.71                                        & 0.65                                            & 0.76               &     \textbf{0.55}                        \\ \cline{2-8} 
                          & Student 003 (CBE)                                  &    1.83                                             &           1.22                                      &        0.96                                         & -                                              & -         &       \textbf{0.72}                              \\ \cline{2-8} 
\multirow{-4}{*}{ADE}    &  gall (VEIIG) & 1.72 & 1.14 & 0.91  & -  & -  & \textbf{0.68}\\ \hline \hline
                         & ETH (BIWI)                                         &        2.20                                         &           1.84                                      &          1.43                                       & 1.52                                            & 1.45             &     \textbf{1.22}                          \\ \cline{2-8} 
                         & Hotel (BIWI)                                       &        2.45                                         &             1.95                                    &             1.65                                    & 1.62                                            & 1.77              &        \textbf{1.43}                      \\ \cline{2-8} 
                       & Student 003 (CBE)                                  &     2.63                                            &           1.97                                      &           1.80                                      & -                                              & -          &        \textbf{1.65}                            \\ \cline{2-8}   
\multirow{-4}{*}{FDE}    & gall (VEIIG) & 2.55 &1.83  & 1.65  & - & - & \textbf{1.45} \\ \hline 
\end{tabular}
}
\end{table}

When observing the results tabulated in Tab. \ref{tab:track_prediction} we observe poor performance for the SF model due to it's lack of capacity to model history. Models So-LSTM and SHA utilise short term history from the pedestrian of interest and the local neighbourhood and generate improved predictions. However we observe a significant increase in performance from methods that optimise generic loss functions such as So-LSTM and SHA to GAN based methods such as So-GAN and SoPhie. This emphasises the need for task specific loss function learning in order to imitate complex human social navigation strategies. In the proposed method we further augment this performance by conditioning the trajectory generator on the proposed neighbourhood encoding mechanism. 

\begin{figure}[htb]
  \centering
    \subfigure[]{\includegraphics[width=4.5cm,height=3.0cm]{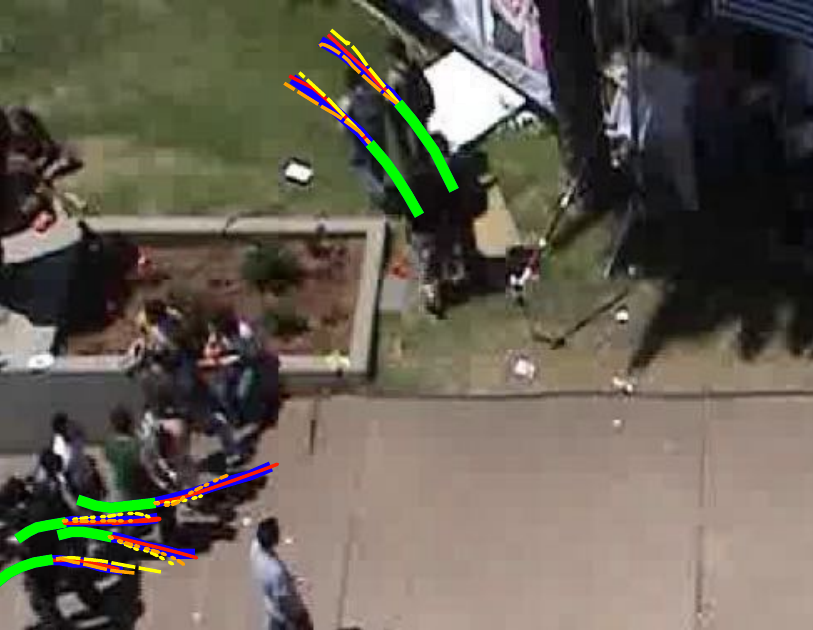}}
    \subfigure[]{\includegraphics[width=4.5cm,height=3.0cm]{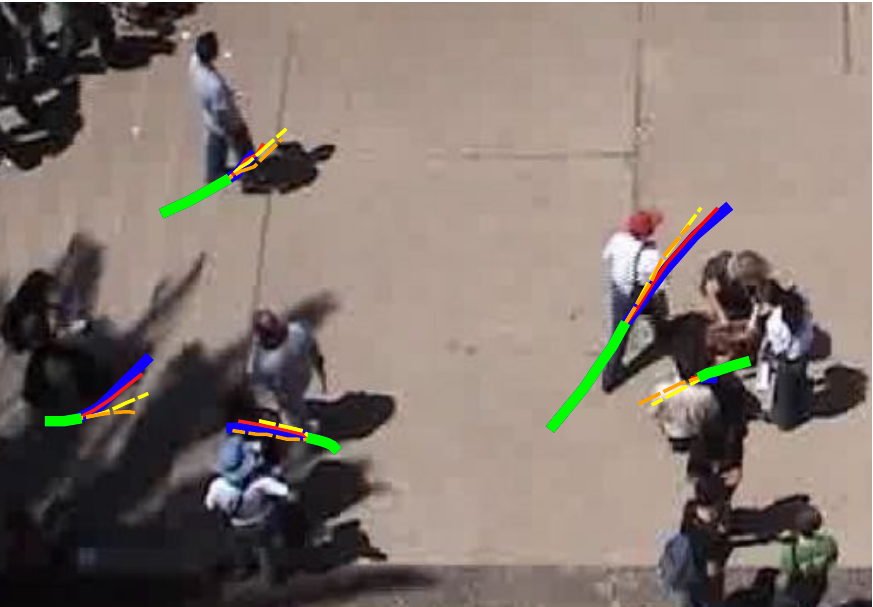}}
    \subfigure[]{\includegraphics[width=4.5cm,height=3.0cm]{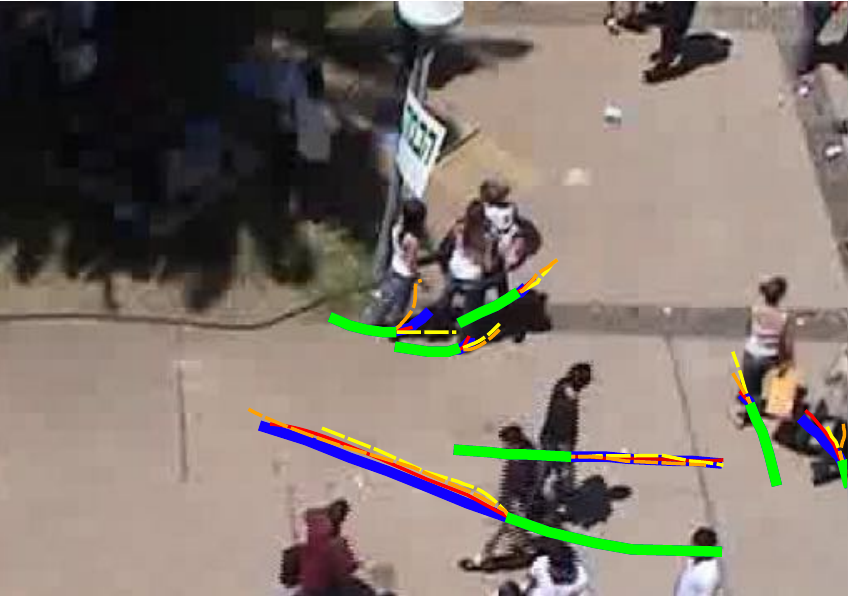}}
    \subfigure[]{\includegraphics[width=4.5cm,height=3.0cm]{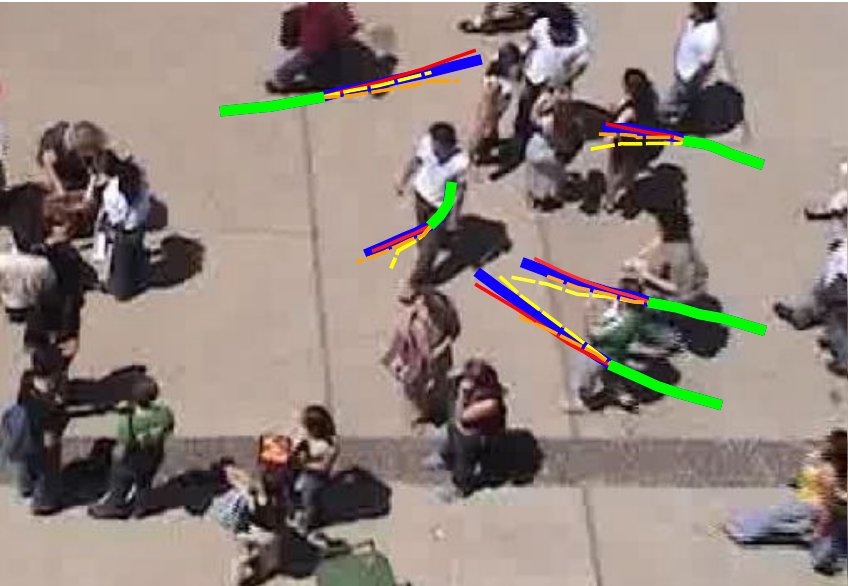}}
   \caption{Qualitative results for the proposed trajectory prediction framework for sequences from the CBE dataset. Given (in green), Ground Truth (in blue) and Predicted trajectories from proposed (in red), SHA model (in yellow) crom So-GAN (in brown). For visual clarity, we show only the trajectories for some of the pedestrians in the scene.}
  \label{fig:track_pred}
\end{figure}

We present a qualitative evaluation of the proposed trajectory generation framework with the SHA and So-GAN baselines in Fig. \ref{fig:track_pred} (selected based on the availability of their implementations). The observed portion of the trajectory is denoted in green, the ground truth observations in blue and predicted trajectories are shown in red (proposed), yellow (SHA) and brown (So-GAN). Observing the qualitative results it can be clearly seen that the proposed model generates better predictions compared to the state-of-the-art considering the varying nature of the neighbourhood clutter. For instance in Fig. \ref{fig:track_pred} (c) and (d) we observe significant deviations between the predictions for SHA and So-GAN and the ground truth. 
However the proposed model better anticipates the pedestrian motion with the improved context modelling and learning process. It should be noted that the proposed method has a better ability to anticipate stationary groups compared to the baselines, which is visible in Fig. \ref{fig:track_pred} (c).

\subsection{Evaluation of the Group Detection}

\subsubsection{Datasets}
Similar to Sec. \ref{sec:ped_eval} we use the BIWI, CBE and VEIIG datasets in our evaluation. Dataset characteristics are reported in Tab. \ref{tab:dataset_charateristics}. 

\begin{table}[htb]
\centering
\caption{Dataset characteristics for different sequences in BIWI \cite{pellegrini2009you}, CBE \cite{lerner2007crowds} and VEIIG \cite{bandini2014towards} datasets}
\label{tab:dataset_charateristics}
\begin{tabular}{|c|c|c|c|c|}
\hline
Dataset    & ETH (BIWI) & Hotel (BIWI) & Student-003 (CBE) & gall (VEIIG) \\ \hline
Frames     & 1448       & 1168         & 541               & 7500        \\ \hline
Pedestrian & 360        & 390          & 434               & 630         \\ \hline
Groups     & 243        & 326          & 288               & 207         \\ \hline
\end{tabular}
\end{table}
\subsubsection{Evaluation Metrics}
One popular measure of clustering accuracy is the pairwise loss $\Delta_{pw}$ \cite{zanotto2012online}, which is defined as the ratio between the number of pairs on which $\beta$ and $\hat{\beta}$ disagree on their cluster membership and the number of all possible pairs of elements in the set. 

However as described in \cite{solera2013structured,solera2016socially} $\Delta_{pw}$ accounts only for positive intra-group relations and neglects singletons. Hence we also measure the Group-MITRE loss, $\Delta_{GM}$, introduced in \cite{solera2013structured}, which has overcome this deficiency. $\Delta_{GM}$ adds a fake counterpart for singletons and each singleton is connected with it's counterpart. Therefore $\delta_{GM}$ also takes singletons into consideration.

\subsubsection{Baselines and Evaluation}
\label{sec:group_detection_eval}
We compare the proposed Group Detection GAN (GD-GAN) framework against 5 recent state-of-the-art baselines, namely \cite{shao2014scene,zanotto2012online,yamaguchi2011you,ge2012vision,solera2016socially}, selected based on their reported performance in public benchmarks. 

In Tab. \ref{tab:group_detection} we report the Precision $(P)$ and Recall $(R)$ values for $\Delta_{pw}$ and $\Delta_{GM}$ for the proposed method along with the state-of-the-art baselines. The proposed GD-GAN method has been able to achieve superior results, especially among unsupervised grouping methods. It should be noted that methods \cite{solera2016socially,shao2014scene,zanotto2012online,yamaguchi2011you} utilise handcrafted features and use supervised learning to separate the groups. As noted in Sec. \ref{sec:intro} these methods cannot adapt to scene variations and require hand labeled datasets for training. Furthermore we would like to point out that the supervised grouping mechanism in \cite{solera2016socially} directly optimises $\Delta_{GM}$. However, without such tedious annotation requirements and learning strategies, the proposed method has been able to generate commendable and consistent results in all considered datasets, especially in cluttered environments \footnote{see the supplementary material for the results for using supervised learning to separate the groups on proposed context features.}. 

\begin{figure}[htb]
  \centering
    \subfigure[GVEII - Frame 2127]{\includegraphics[width=.4\linewidth]{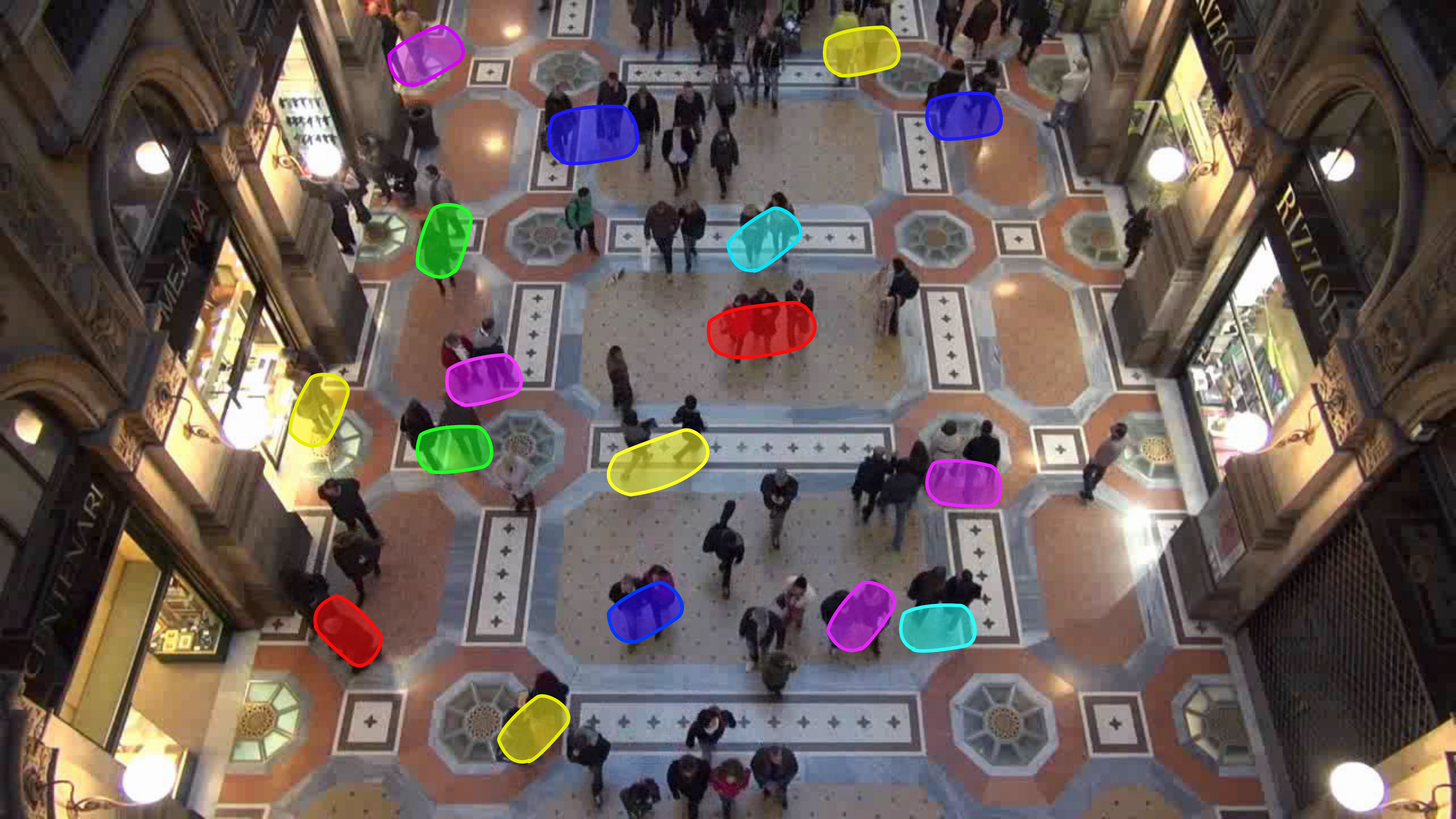}}
    \subfigure[GVEII- Frame 2320]{\includegraphics[width=.4\linewidth]{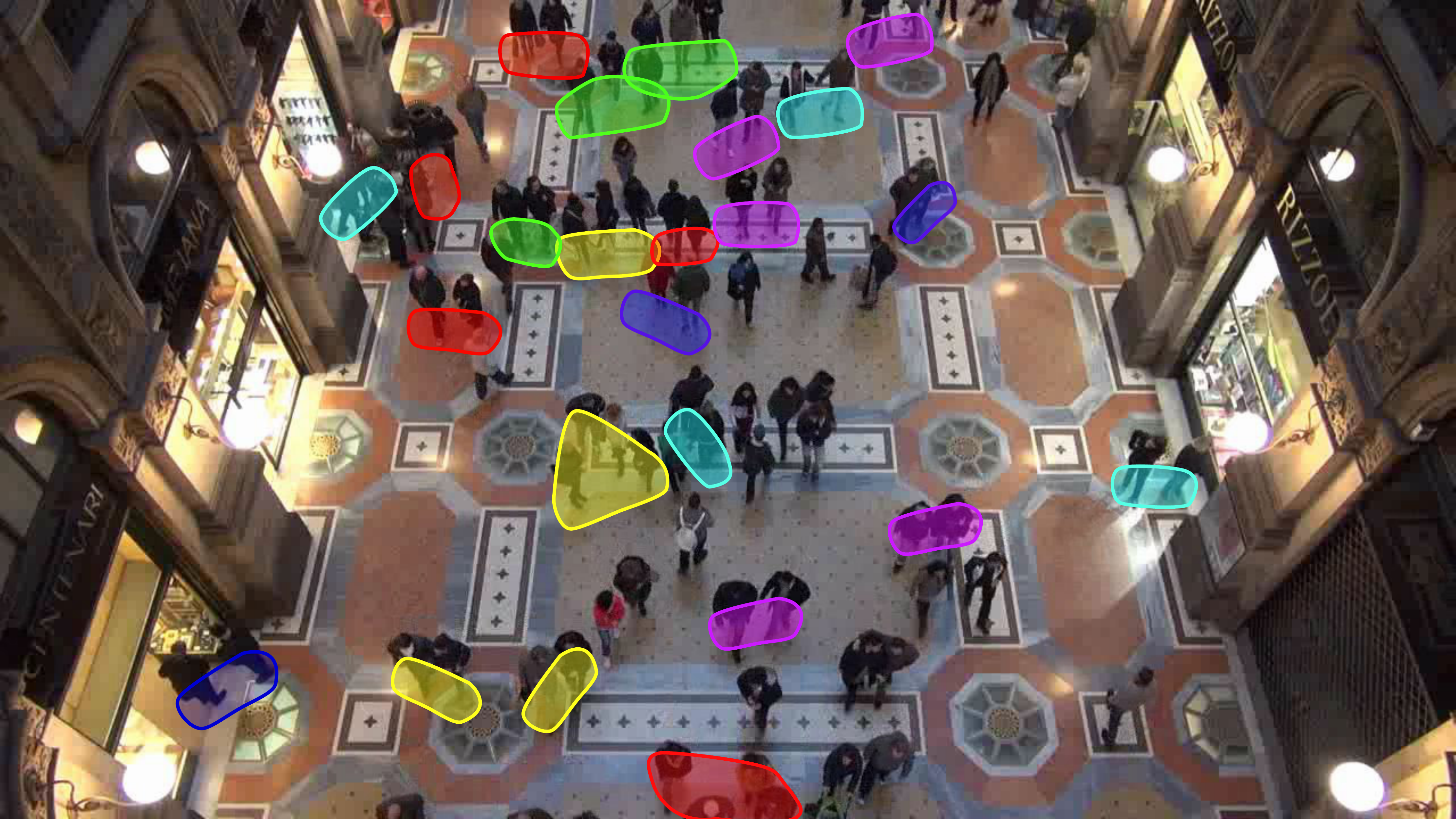}}
    \subfigure[CBE - Frame 2603]{\includegraphics[width=.4\linewidth]{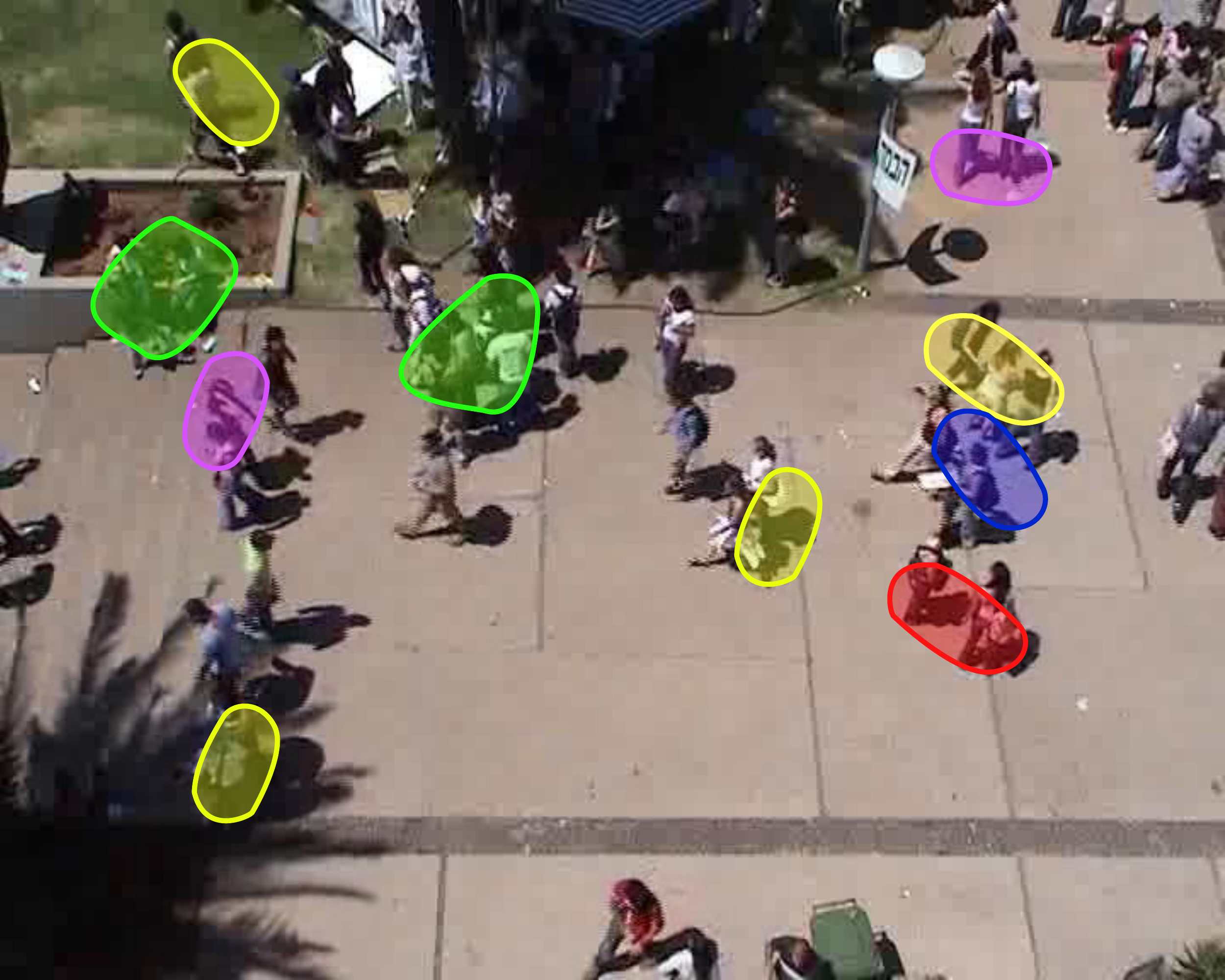}}
    \subfigure[CBE - Frame 2910]{\includegraphics[width=.4\linewidth]{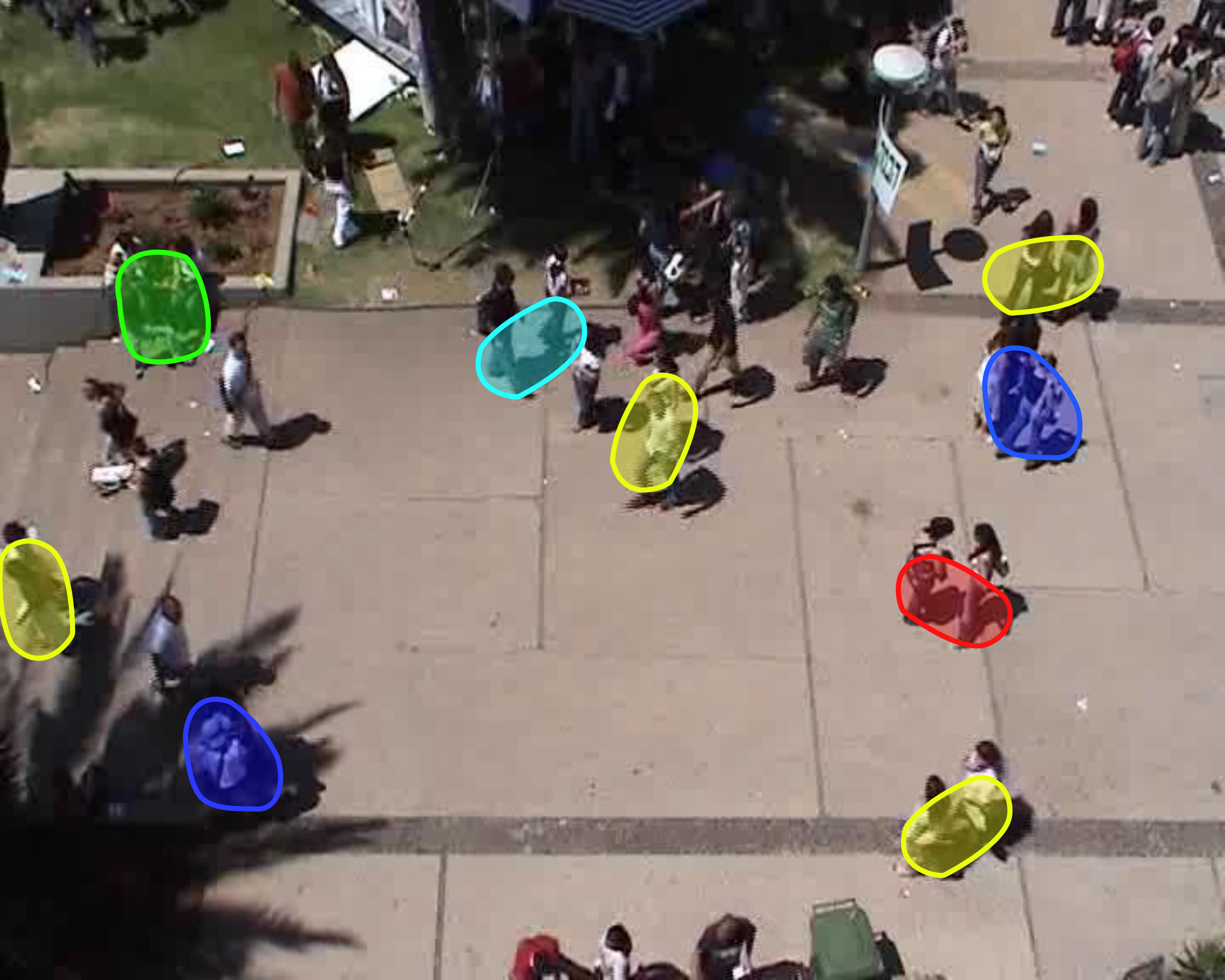}}
   \caption{Qualitative results from the proposed GD-GAN methods for sequences from the CBE and GVEII datasets. Connected coloured blobs indicate groups of pedestrians.}
  \label{fig:group_pred}
\end{figure}

In Fig. \ref{fig:group_pred} we show groups detected by the proposed GD-GAN method for sequences from the CBE and VEIIG datasets. Regardless of the scene context, occlusions and the varying crowd densities, the proposed GD-GAN method generates acceptable results. We believe this is due to the augmented features that we derive through the automated deep feature learning process. These features account for both historical and future behaviour of the individual pedestrians, hence possessing an ability to detect groups even in the presence of occlusions such as in Fig. \ref{fig:group_pred} (c). 

\begin{table}[htb]
\centering
\caption{Comparative results on the BIWI \cite{pellegrini2009you}, CBE \cite{lerner2007crowds} and VEIIG \cite{bandini2014towards} datasets using the $\Delta_{GM}$ \cite{solera2013structured} and $\Delta_{PW}$ \cite{zanotto2012online} metrics. `-' refers to unavailability of that specific evaluation. The best results are shown in bold and the second best results are underlined.}
\label{tab:group_detection}
\resizebox{1.0\textwidth}{!}{
\begin{tabular}{|l|>{\centering}m{1cm}|>{\centering}m{1cm}|>{\centering}m{1.25cm}|>{\centering}m{1.25cm}|>{\centering}m{1.36cm}|>{\centering}m{1.36cm}|>{\centering}m{1.2cm}|>{\centering}m{1.2cm}|>{\centering}m{1.2cm}|>{\centering}m{1.2cm}|>{\centering}m{1.2cm}|m{1.2cm}|}
\hline
\multicolumn{1}{|c|}{}                                                           & \multicolumn{2}{c|}{Shao et. al \cite{shao2014scene}}                                  & \multicolumn{2}{c|}{zanotto et. al \cite{zanotto2012online}}                               & \multicolumn{2}{c|}{ Yamaguchi et. al \cite{yamaguchi2011you}}                                  & \multicolumn{2}{c|}{Ge et. al \cite{ge2012vision}} & \multicolumn{2}{c|}{Solera et al. \cite{solera2016socially}} & \multicolumn{2}{c|}{GD-GAN} \\ \cline{2-13} 
\multicolumn{1}{|c|}{\multirow{-2}{*}{}}                                         & $P$                                                 & $R$    & $P$                                              & $R$    & $P $                                                & $R $   & $P$          & $R$         & $P $          & $R$           & $P$            & \hspace{.32cm} $R   $         \\ \hline
\rowcolor[HTML]{EFEFEF} 
BIWI  \hspace*{1.2cm} $\Delta_{GM}$ & 67.3 & 64.1 & - & -    & 84.0 & 51.2 & 89.2       & 90.9      & \underline{97.3}        & \textbf{97.7}        &   \textbf{97.5}         &\hspace{.2cm} \textbf{97.7}            \\
Hotel   \hspace*{1.2cm}  $\Delta_{PW}$                        & 51.5                         & 90.4 & 81.0                      & 91.0  &83.7                         & \textbf{93.9} & 88.9       & 89.3      & \underline{89.1}        & 91.9        &  \textbf{90.2}           &\hspace{.2cm} \underline{93.1}            \\ \hline
\rowcolor[HTML]{EFEFEF} 
BIWI \hspace*{1.2cm} $\Delta_{GM}$ & 69.3 & 68.2 & - & -    & 60.6 & 76.4 & 87.0       & 84.2      & \underline{91.8}        & \textbf{94.2}        &    \textbf{92.5}          &   \hspace{.2cm} \textbf{94.2}          \\
ETH \hspace*{1.2cm} $\Delta_{PW}$                                                & 44.5                                              & 87.0 & 79.0                                           & 82.0 & 72.9                                              & 78.0 & 80.7       & 80.7      & \underline{91.1}        & \underline{83.4}        &   \textbf{91.3}           &  \hspace{.2cm} \textbf{83.5}            \\ \hline
\rowcolor[HTML]{EFEFEF} 
CEB \hspace*{1.4cm}$\Delta_{GM}$                                                & 40.4                                              & 48.6 & -                                              & -    & 56.7                                              & 76.0 & 77.2       & 73.6      & \textbf{81.7}        & \textbf{82.5}        &      \underline{81.0}         &  \hspace{.2cm} \underline{81.8}            \\
Student-003 \hspace*{0.4cm}$\Delta_{PW}$                                        & 10.6                                              & \textbf{76.0} & 70.0                                           & 74.0 & 63.9                                              & 72.6 & 72.2       & 65.1      & \textbf{82.3}        & \underline{74.1}        &   \underline{82.1}           &       \hspace{.2cm} 63.4       \\ \hline
\rowcolor[HTML]{EFEFEF} 
VEIIG \hspace*{1.1cm}$\Delta_{GM}$                                              &                        -                           &     - &                -                                &  -    &             -                                      &   -   &     -       &        -   &  \textbf{84.1}           &   \textbf{84.1}          & \underline{83.1}             &    \hspace{.2cm} \underline{79.5}          \\
gall \hspace*{1.5cm}$\Delta_{PW}$                                               &          -                                         &  -    &               -                                 &  -    &               -                                    &  -    &      -      &   -        &     \textbf{79.7}        &   \textbf{77.5}          &   \underline{77.6}           &  \hspace{.2cm}  \underline{73.1}           \\ \hline
\end{tabular}
}
\end{table}
 
\begin{figure}[htb]
  \centering
    \includegraphics[width=0.9\linewidth]{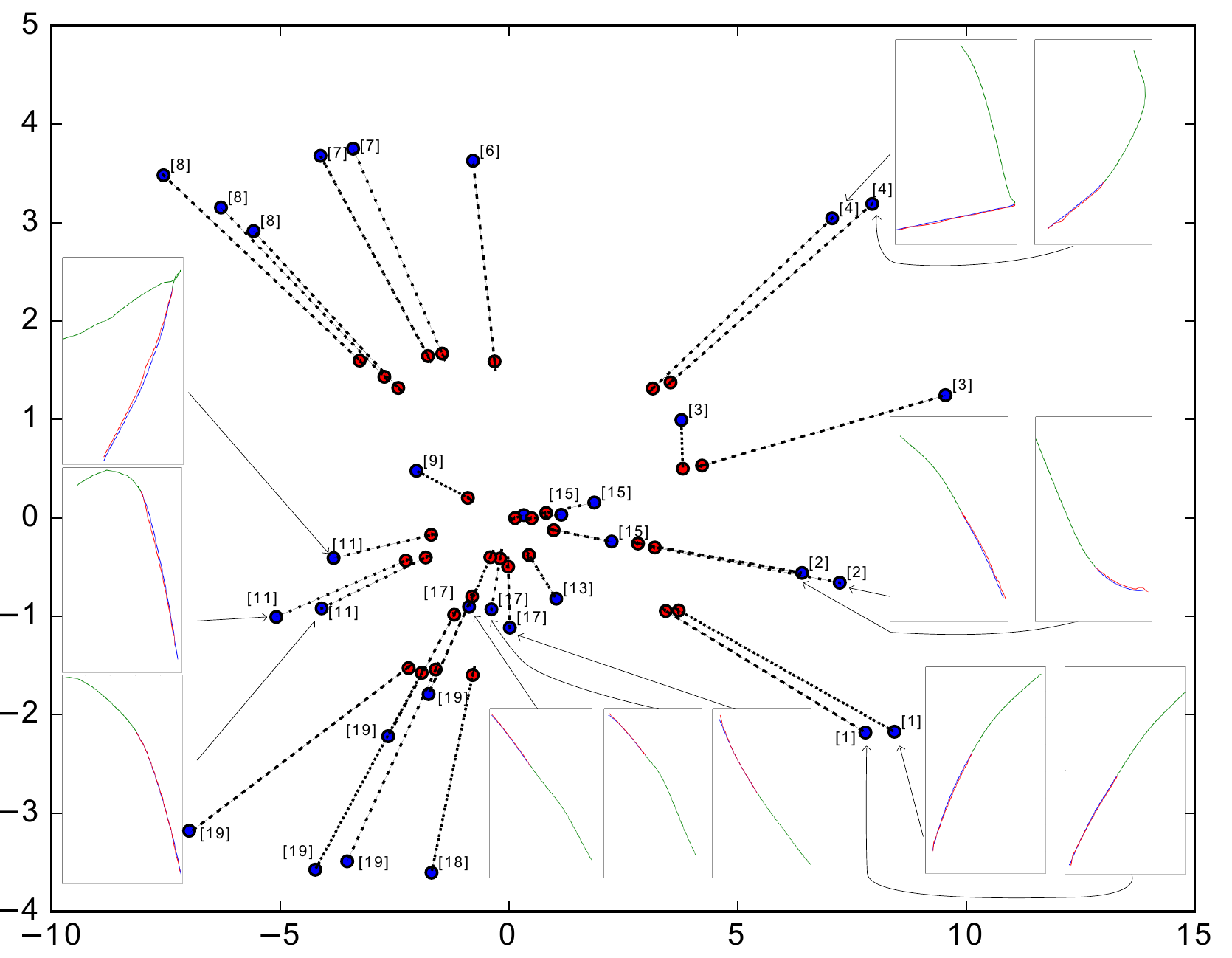}
   \caption{Projections of the trajectory generator (G) hidden states before (in blue) and after (in red) training. Ground truth group IDs are in brackets. Each insert indicates the trajectory associated with the embedding. The given portion of the trajectory is in green, and the ground truth and prediction are in blue and red respectively}
  \label{fig:embedding_shift}
\end{figure}

We selected the first 30 pedestrian trajectories from the VEIIG test set and in Fig. \ref{fig:embedding_shift} we visualise the embedding space positions before (in blue) and after (in red) training of the proposed trajectory generator (G). Similar to \cite{aubakirova2016interpreting} we extracted the activations using the feature extractor function $f$ and applied PCA \cite{wold1987principal} to plot them in 2D. The respective ground truth group IDs are indicated in brackets. This helps us to gain an insight into the encoding process that $G$ utilises, which allows us to discover groups of pedestrians. Considering the examples given, it can be seen that trajectories from the same cluster become more tightly grouped. This is due to the model incorporating source positions, heading direction, trajectory similarity, when embedding trajectories, allowing us to extract pedestrian groups in an unsupervised manner. 

\subsection{Ablation Experiment}
To further demonstrate the proposed group detection approach, we conducted a series of ablation experiments identifying the crucial components of the proposed methodology~\footnote{see the supplementary material for an ablation study for the trajectory prediction}. In the same setting as the previous experiment we compare the proposed GD-GAN model against a series of counter parts as follows:  

\begin{itemize}
\item GD-GAN / GAN: removes $D$ and the model $G$ is learnt through supervised learning as in \cite{fernando2017soft+}. 
\item GD-GAN / cGAN: optimises the generic GAN objective defined in \cite{goodfellow2014generative}.  
\item GD-GAN / $L_1$: removes sparsity regularisation and optimises Eq. \ref{eq:gan}.
\item GD-GAN + hf: utilises features from $G$ as well as the handcrafted features defined in \cite{solera2016socially} for clustering.
\end{itemize}

\begin{table}[htbp]
\centering
\caption{Ablation experiment evaluations}
\label{tab:ablation_experiment}
\resizebox{1.0\textwidth}{!}{
\begin{tabular}{| l |>{\centering}m{1.5cm}| >{\centering}m{1.5cm} |>{\centering}m{1.5cm}| >{\centering}m{1.5cm} |>{\centering}m{1.5cm}| >{\centering}m{1.5cm} |>{\centering}m{1.5cm}| >{\centering}m{1.5cm} |>{\centering}m{1.5cm}| m{1.5cm} |}
\hline
                                          & \multicolumn{2}{>{\centering}m{3cm} |}{GD-GAN / GAN} & \multicolumn{2}{>{\centering}m{3cm} |}{GD-GAN / cGAN} & \multicolumn{2}{>{\centering}m{3cm} |}{GD-GAN / $L_1$} & \multicolumn{2}{>{\centering}m{3cm} |}{GD-GAN + hf} & \multicolumn{2}{>{\centering}m{3cm} |}{GD-GAN} \\ \cline{2-11} 
\multirow{-2}{*}{}                        & P               & R               & P                & R               & P                & R                & P               & R              & P            & \hspace{.6cm}R            \\ \hline
\rowcolor[HTML]{EFEFEF} 
CEB \hspace{0.98 cm} $\Delta_{GM}$         & 73.6            & 75.1            & 76.7             & 76.2            & 77.3             & 78.0             & \textbf{79.0}            & \textbf{79.2}           & 78.7         & \hspace{.5cm}\textbf{79.2}         \\
Student-003 $\Delta_{PW}$ & 74.1            & 52.8            & 75.5             & 60.2            & 78.1             & 65.1             & \textbf{80.4}            & 68.0           & \textbf{80.4}         & \hspace{.5cm}\textbf{68.4}         \\ \hline
\end{tabular}
}
\end{table}

The results of our ablation experiment are presented in Tab. \ref{tab:ablation_experiment}. Model GD-GAN / GAN performs poorly due to the deficiencies in the supervised learning process. It optimises a generic mean square error loss, which is not ideal to guide the model through the learning process when modelling a complex behaviour such as human navigation. Therefore the resultant feature vectors do not capture the full context which contributes to the poor group detection accuracies. We observe an improvement in performance with GD-GAN / cGAN due to the GAN learning process which is further augmented and improved through GD-GAN / $L_1$ where the model learns a conditional behaviour depending on the neighbourhood context. $L_1$ regularisation further assists the group detection process via making the learnt feature distribution more discriminative. 

In order to demonstrate the credibility of the learnt group attributes from the proposed GD-GAN model, we augment the feature vector extracted in Eq. \ref{eq:feature_extractor} together with the features proposed in \cite{solera2016socially} and apply subsequent process (i.e Eq. \ref{eq:t_SNE} and \ref{eq:DBSCAN}) to discover the groups. We utilise the public implementation \footnote{https://github.com/francescosolera/group-detection} released by the authors for the feature extraction.

We do not observe a substantial improvement with the group detection performance being very similar, indicating that the proposed GD-GAN model is sufficient for modelling the social navigation structure of the crowd. 

\subsection{Time efficiency}
We use the Keras \cite{chollet2017keras} deep learning library for our implementation. The GD-GAN module does not require any special hardware such as GPUs to run and has 41.8K trainable parameters. We ran the test set in Sec. \ref{sec:group_detection_eval} on a single core of an Intel Xeon E5-2680 2.50GHz CPU and the GD-GAN algorithm was able to generate 100 predicted trajectories with 30, 2 dimensional data points in each trajectory (i.e. using 15 observations to predict the next 15 data points) and complete the group detection process in 0.712 seconds.

\section{Conclusions}
In this paper we have proposed an unsupervised learning approach for pedestrian group segmentation. We avoid the the need to handcraft sociological features by automatically learning group attributes through the proposed trajectory prediction framework. This allows us to discover a latent representation accounting for both historical and future behaviour of each pedestrian, yielding a more efficient platform for detecting their social identities. Furthermore, the unsupervised learning setting grants the approach the ability to employ the proposed framework in different surveillance settings without tedious learning of group memberships from a hand labeled dataset. Our quantitative and qualitative evaluations on multiple public benchmarks clearly emphasise the capacity of the proposed GD-GAN method to learn complex real world human navigation behaviour. 


\end{document}